\pdfoutput=1

\documentclass[11pt]{article}

\usepackage[final]{acl}

\usepackage{times}
\usepackage{latexsym}

\usepackage[T1]{fontenc}

\usepackage[utf8]{inputenc}

\usepackage{microtype}

\usepackage{inconsolata}

\usepackage{graphicx}
\usepackage{booktabs}
\usepackage{multirow}
\usepackage{comment}

\title{SeaLLMs 3: Open Foundation and Chat Multilingual Large Language Models for Southeast Asian Languages}

\author{%
    Wenxuan Zhang\thanks{Equal contributions.}, \ Hou Pong Chan$^*$, Yiran Zhao$^*$, Mahani Aljunied$^*$, Jianyu Wang$^*$, \\
    \textbf{Chaoqun Liu, Yue Deng, Zhiqiang Hu, Weiwen Xu, Yew Ken Chia, Xin Li, Lidong Bing\thanks{ Corresponding author: \href{mailto:l.bing@alibaba-inc.com}{l.bing@alibaba-inc.com}}}  \\ \\
  DAMO Academy, Alibaba Group \\ \\
  Project page: \url{https://seallms.github.io/}
}

\begin{document}
\maketitle

\begin{abstract}
Large Language Models (LLMs) have shown remarkable abilities across various tasks, yet their development has predominantly centered on high-resource languages like English and Chinese, leaving low-resource languages underserved. To address this disparity, we present SeaLLMs 3, the latest iteration of the SeaLLMs model family, tailored for Southeast Asian languages. This region, characterized by its rich linguistic diversity, has lacked adequate language technology support. SeaLLMs 3 aims to bridge this gap by covering a comprehensive range of languages spoken in this region, including English, Chinese, Indonesian, Vietnamese, Thai, Tagalog, Malay, Burmese, Khmer, Lao, Tamil, and Javanese. Leveraging efficient language enhancement techniques and a specially constructed instruction tuning dataset, SeaLLMs 3 significantly reduces training costs while maintaining high performance and versatility. Our model excels in tasks such as world knowledge, mathematical reasoning, translation, and instruction following, achieving state-of-the-art performance among similarly sized models. Additionally, we prioritized safety and reliability by addressing both general and culture-specific considerations and incorporated mechanisms to reduce hallucinations. This work underscores the importance of inclusive AI, showing that advanced LLM capabilities can benefit underserved linguistic and cultural communities.
\end{abstract}

\section{Introduction}
Large Language Models (LLMs) such as GPT-4 \cite{gpt-4} and Gemini \cite{gemini} have demonstrated remarkable capabilities across a wide array of tasks, ranging from natural language understanding and generation to more specialized domain applications \cite{llm-survey-zhaoxin}. These models have proven valuable, offering substantial benefits to the global community, especially through the proliferation of open-source LLMs such as Llama \cite{llama-1, llama-2}, Mistral \cite{mistral-7b}, Qwen \cite{qwen-1, yang2024qwen2}, and Gemma \cite{gemma-1}.
However, the majority of these efforts have been concentrated on high-resource languages such as English and Chinese, or well-developed regions like Europe \cite{m3exam, multilingual-eval-emnlp23}. Consequently, the development of LLMs tailored for low-resource languages or underdeveloped regions has been significantly overlooked, resulting in a lack of inclusivity and equitable distribution of AI advancements across diverse linguistic and cultural communities \cite{multilingual-survey-libo, multilingual-survey-kaiyu, translation-analysis}.

To bridge this gap, we introduced the SeaLLMs model \cite{seallms}, specifically designed LLMs for Southeast Asian languages. Southeast Asia (SEA) is a region with a rich diversity of languages spoken by millions of people, yet it suffers from a significant lack of language technology support \cite{acl22-indonlp}. 
The initiative of SeaLLMs thus aims to make the benefits of LLMs accessible to speakers of these languages, addressing their unique linguistic and cultural nuances. 
Following this endeavor, several other models have been dedicated to this region as well, such as SEA-LION \cite{sea_lion_2023} and Sailor \cite{sailor}. 
However, these models often face significant limitations: they are typically released only as foundational or chat models, offer limited options in terms of model size, and cover a limited number of SEA languages. Moreover, the relatively scarce availability of language corpora further constrains the amount of training data available, hindering the development and performance of these models.

In this work, we introduce \textbf{SeaLLMs 3}, the latest iteration of the SeaLLMs model family. This version is designed to cover a more diverse array of Southeast Asian languages, including English, Chinese, Indonesian, Vietnamese, Thai, Tagalog, Malay, Burmese, Khmer, Lao, Tamil, and Javanese. 
Different from the conventional continue-pretraining paradigm \cite{language-enhancement-llama, seallms, sailor}, we conduct efficient language enhancement by training language-specific neurons only based on a foundation model \cite{multilingual-analysis}, significantly reducing the overall training cost. Moreover, such targeted training also ensures that the performance of high-resource languages can remain unaffected during the enhancement. 
Furthermore, SeaLLMs 3 is trained using a specially constructed instruction tuning dataset that encompasses a wide variety of task types and carefully balanced language distributions. This approach ensures that the model can handle the linguistic diversity of the Southeast Asian region while maintaining high performance and versatility across different applications.
As a result, it achieves state-of-the-art performance among models with similar sizes, excelling across a diverse array of tasks such as world knowledge \cite{m3exam}, mathematical reasoning \cite{mgsm-dataset}, translation \cite{flores-dataset}, and instruction following.

In the meantime, we pay special attention to the model's reliability and trustworthiness during its development, which are often under-considered in multilingual settings \cite{multilingual-jailbreak}. In particular, we address both general and culture-specific safety considerations to ensure the models provide contextually appropriate responses. The model is also specifically trained to be aware of its knowledge boundary and refuse what it does not know. To evaluate this capability, we introduce a novel benchmark, SeaRefuse, which assesses the ability of LLMs to refuse questions beyond their knowledge boundaries. This focus on safety and reliability has resulted in SeaLLMs 3 exhibiting reduced hallucination and delivering safe, coherent responses, especially in queries closely related to Southeast Asian culture.

We open-source both the foundational and chat models of SeaLLMs 3\footnote{\url{https://huggingface.co/collections/SeaLLMs/seallms-v3-668f3a52e1e6fbaad5752cdb}}. The foundational model could serve as a base for conducting instruction tuning tailored to specific application requirements. Meanwhile, the chat model has already undergone instruction tuning and is ready for direct use of handling a wide range of tasks.

\begin{figure}[t]
  \includegraphics[width=\columnwidth]{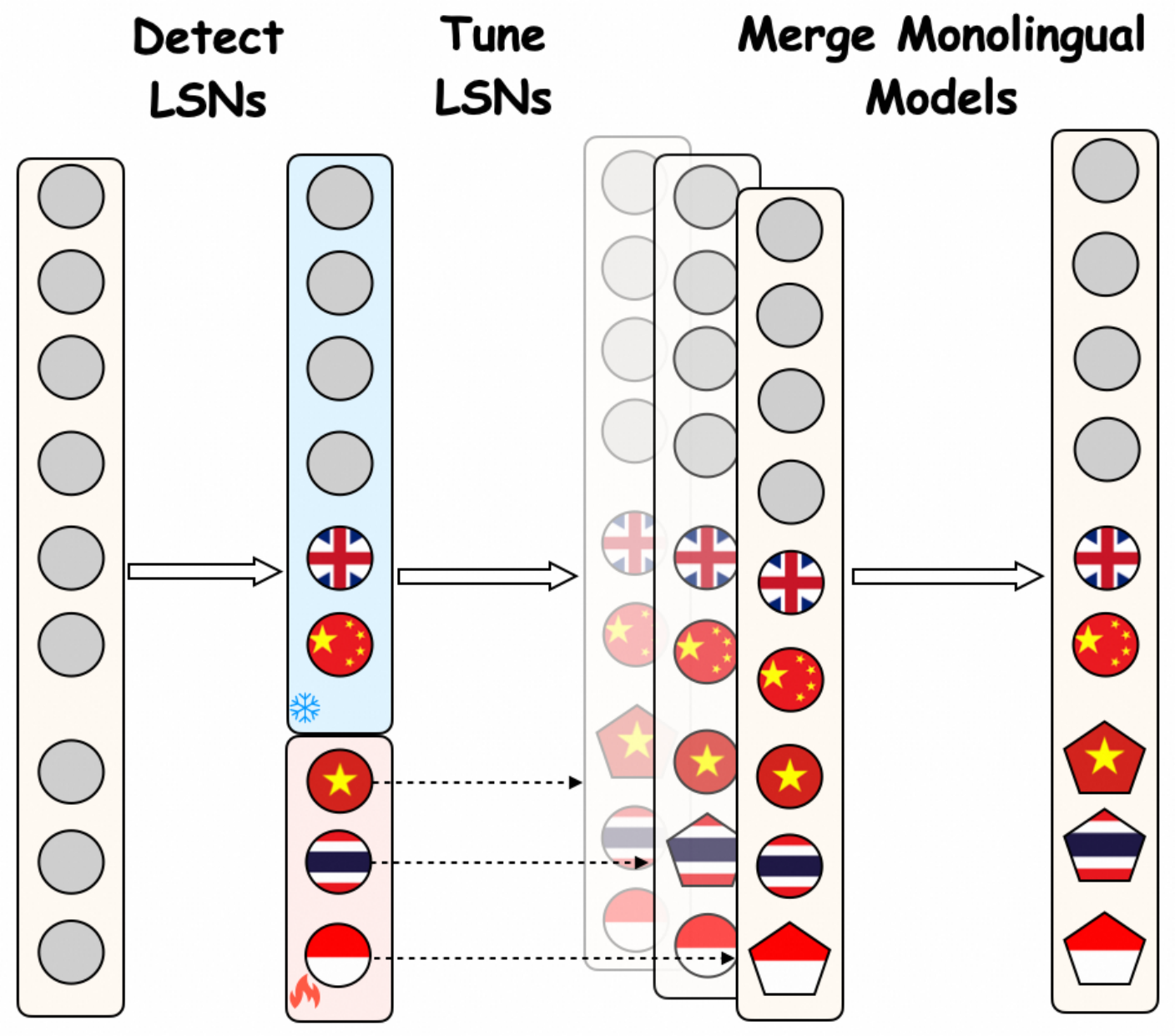}
  \caption{Language-Specific Neuron Training.}
  \label{fig:pretrain}
\end{figure}

\section{Pre-training}

\subsection{Pre-training Data}
Building on the efforts from previous versions of SeaLLMs \cite{seallms}, we have incorporated corpora from a wider range of data sources to enhance diversity.
Specifically, we have integrated fundamental knowledge from Wikipedia \citep{wikidump} and textbooks \citep{benallal2024cosmopedia}, journalistic materials such as CC-News \citep{ccnews}, web-based corpora from CulturaX \citep{nguyen2023culturax}, and MADLAD-400 \citep{kudugunta2023madlad400}. 
We have also improved the data processing pipeline including the language model filtering and duplicate removal to improve the data quality.

Additionally, we explore the utilization of model-synthetic data for training, which received much attention recently \cite{benallal2024cosmopedia}. Starting with manual annotation of domain-specific knowledge points in SEA languages, we then employed stronger models to generate targeted tutorial-style content, thereby enhancing SeaLLMs 3 with enriched regional knowledge in a more explicit form.

\subsection{Language-Specific Neuron Training}
We built our model based on the Qwen2 model family \cite{yang2024qwen2} and further conducted language enhancement to augment its capability in SEA languages. This approach allows the model to quickly inherit foundational knowledge from Qwen, rather than learning it from scratch.

The most straightforward method for language enhancement is typically through continued pertaining \cite{language-enhancement-llama}, which we also used for previous versions of SeaLLMs. However, as discussed, the relatively scarce availability of language corpora limits the amount of training data, hindering the development and performance of these models. Furthermore, it is often observed that such direct continued pretraining can compromise the model's original capacity in high-resource languages like English and Chinese \cite{sailor}.

In this iteration, we adopt Language-Specific Neuron (LSN) training for efficient language enhancement, as shown in Figure \ref{fig:pretrain}. Recent studies have found that certain language-specific neurons in language models are responsible for processing specific languages. For instance, \citet{multilingual-analysis} discovered that language-specific neurons comprise only about 0.1\% of all parameters. Thus, the capabilities of a language can be enhanced by training its corresponding LSNs while preserving the multilingual abilities of other languages.
To efficiently train SeaLLMs 3, we employ the parallel language-specific neuron detection method proposed by \citet{multilingual-analysis}. As shown in the left most part of Figure \ref{fig:pretrain}, this method allows us to identify the LSNs of SEA languages using language-specific training data, selected as a downsampled subset of the corpora from the training data. We then specifically train these detected LSNs to develop multiple monolingual LLMs in SEA languages, which are subsequently merged to create a unified multilingual LLM for SEA languages. Additionally, to maintain proficiency in English and Chinese from the original foundation model, we detect their respective LSNs and exclude them from the entire pre-training process.

This method offers several advantages. First, it requires relatively less training data since the training is more targeted, which significantly reduces training costs. Second, because the training is targeted, we can ensure that the performance of high-resource languages from the original foundation model remains unaffected. LSNs operate independently and do not influence one another, avoiding the sacrifices seen with previous methods.

\section{Supervised Fine-Tuning (SFT)}
\subsection{Supervised Fine-tuning Data}

Most existing open-source supervised fine-tuning (SFT) datasets are predominantly in English \cite{flan, alpaca}, which presents a challenge for developing effective models for Southeast Asian languages. To address this, we employed various techniques to construct our SFT pool. For example, we selectively translate some high-quality English data to SEA languages with quality filtering, conduct self-instruction to automatically generate SFT data of certain types, and use various prompting strategies \cite{nips23-refine, acl24-phi}.
Following our previous practice, native speakers have been actively engaged throughout the entire SFT data construction process. They manually collect and write seed questions and topic lists, ensuring linguistic and cultural accuracy from the outset. Additionally, native speakers verify, filter, and edit the synthetic SFT data to maintain high quality. 

\begin{figure}[t]
  \includegraphics[width=\columnwidth]{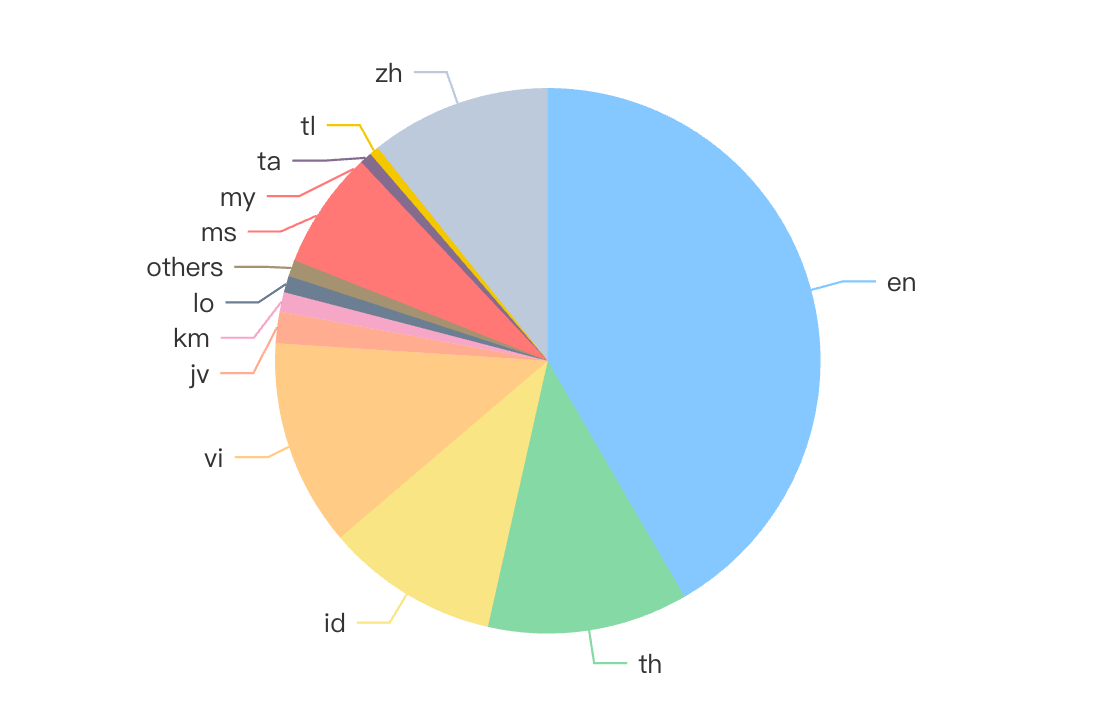}
  \caption{Language distribution of the SFT data}
  \label{fig:sft-language}
\end{figure}

Our preliminary experiments indicated that relying heavily on dominant English data adversely affects performance. To mitigate this, we strive to maintain a relatively good balance of language representation in our training data this time. Figure \ref{fig:sft-language} shows the language distribution of our SFT data. While English remains a significant portion of the dataset, substantial representation is given to other Southeast Asian languages such as Indonesian, Vietnamese, Thai, and others, ensuring a comprehensive and diverse linguistic foundation for the model's training.

Since the first release of SeaLLMs \cite{seallms}, the task types of SFT data have been significantly expanded. The dataset now includes a diverse range of task types such as coding, math, education-related content, reasoning, general dialogue, table-related tasks, open-domain QA, and many more. This expansion ensures that the model is well-rounded and capable of handling a variety of queries and tasks. Additionally, SFT with multiple turns has been significantly increased to enhance the model's ability to engage in natural, multi-turn dialogues, improving its conversational fluency and coherence.

\begin{table*}[th]
    \centering
    \resizebox{0.75\textwidth}{!}{
    \begin{tabular}{lccccccc}
    \toprule
    \textbf{\normalsize{Model}} & \textbf{\normalsize{en}} & \textbf{\normalsize{zh}} & \textbf{\normalsize{id}} & \textbf{\normalsize{th}} & \textbf{\normalsize{vi}} & \textbf{\normalsize{avg}} & \textbf{\normalsize{avg\_sea}} \\
    \midrule
    Gemma-7B & 73.2 & 51.9 & 47.5 & 46.0 & 59.4 & 55.6 & 51.0 \\
    Sailor-7B-Chat & 66.0 & 65.2 & 47.5 & 46.2 & 51.3 & 55.2 & 48.3 \\
    SeaLLM-7B-v2.5 & 75.8 & 58.1 & 49.9 & 50.2 & 62.2 & 59.2 & 54.1 \\
    Sailor-14B & 74.8 & 84.0 & 53.6 & 52.8 & 62.1 & 65.5 & 56.2 \\
    Sailor-14B-Chat & 74.9 & 84.3 & 55.3 & 56.6 & 63.7 & 67.0 & 58.5 \\
    Qwen2-7B & \textbf{81.5} & 87.4 & 53.0 & 47.9 & 62.8 & 66.5 & 54.6 \\
    Qwen2-7B-Instruct & 80.9 & \textbf{88.0} & 55.8 & 55.5 & 62.4 & 68.5 & 57.9 \\
    \midrule
    \textbf{SeaLLMs-v3-7B} & 80.9 & 86.3 & 54.5 & 53.0 & 62.8 & 67.5 & 56.8 \\
    \textbf{SeaLLMs-v3-7B-Chat} & 80.9 & 87.4 & \textbf{55.8} & \textbf{56.9} & \textbf{64.9} & \textbf{69.2} & \textbf{59.2} \\

    \bottomrule
    \end{tabular}
    }
    \caption{Results of multilingual world knowledge with the M3Exam benchmark}
    \label{tab:world-knowledge}
\end{table*}

Model safety, trustworthiness, and reliability are also important factors for constructing the SFT pool. To address this, we specifically constructed refusal-type data, enabling the model to decline questions beyond its knowledge boundaries, such as those involving non-existing entities. Furthermore, we carefully curated safety-related data, including both general safety data (which are culturally independent, such as general moral principles) and country-specific safety data (which are culturally sensitive). This approach ensures that the model can be safely deployed with cultural considerations in mind, providing accurate and appropriate responses across different cultural contexts.

\subsection{Training Details}
Two stages of training are employed to optimize the model's performance. In the first stage, a large volume of SFT data is used to equip the model with instruction-following capabilities and to familiarize it with different task types. In the second stage, a smaller but high-quality SFT dataset is utilized to fine-tune the model, ensuring it performs exceptionally well on important tasks.

During the training process, different samples are packed together for efficiency, with a maximum length set at 8,192 tokens. The learning rate is set at 1.0e-5, with a warmup ratio of 0.1. Additionally, gradients are clipped at a maximum of 1.0 to prevent exploding gradients.

\begin{table*}[th]
    \centering
    \resizebox{0.75\textwidth}{!}{
    \begin{tabular}{lccccccc}
    \toprule
    \textbf{\normalsize{Model}} & \textbf{\normalsize{en}} & \textbf{\normalsize{zh}} & \textbf{\normalsize{id}} & \textbf{\normalsize{th}} & \textbf{\normalsize{vi}} & \textbf{\normalsize{avg}} & \textbf{\normalsize{avg\_sea}} \\
    \midrule
    Gemma-7B & 63.4 & 50.9 & 54.5 & 49.0 & 49.4 & 53.5 & 51.0  \\
    Sailor-7B-Chat & 55.8 & 47.2 & 48.4 & 41.4 & 46.2 & 47.8 & 45.4  \\
    SeaLLM-7B-v2.5 & 65.2 & 54.4 & 56.5 & 47.9 & 52.8 & 55.3 & 52.4   \\
    Sailor-14B & 61.8 & 56.4 & 57.0 & 48.2 & 53.5 & 55.4 & 52.9  \\
    Sailor-14B-Chat & 62.7 & 56.1 & 56.7 & 49.6 & 54.1 & 55.8 & 53.5 \\
    Qwen2-7B & 71.0 & 64.2 & 60.2 & 52.0 & 56.6 & 60.8 & 56.3  \\
    Qwen2-7B-Instruct & 70.8 & 63.5 & 59.9 & 52.4 & 56.8 & 60.7 & 56.4  \\
    \midrule
    \textbf{SeaLLMs-v3-7B} & 70.6 & \textbf{65.4} & 61.7 & 53.6 & \textbf{58.7}  & 62.0 & 58.0 \\
    \textbf{SeaLLMs-v3-7B-Chat} & \textbf{71.3} & 64.7 & \textbf{62.5} & \textbf{54.4} & 57.8 & \textbf{62.2} & \textbf{58.2} \\

    \bottomrule
    \end{tabular}
    }
    \caption{Results of multilingual world knowledge with the translated MMLU benchmark}
    \label{tab:mmlu}
\end{table*}

\section{Evaluations}
We conduct extensive evaluations against models with similar sizes, including Sailor-7B / Sailor-7B-Chat \cite{sailor}, Gemma-7b / Gemma-7b-it \cite{gemma-1}, Qwen2-7B / Qwen2-7B-Chat \cite{yang2024qwen2}, Meta-Llama-3-8B / Meta-Llama-3-8B-Instruct \cite{llama-2}, Aya-23-8B \cite{aya-model}, and the previous versions (mainly v2.5) of the SeaLLMs \cite{seallms}. The models are listed by their release date in the following result tables.

The evaluations can be generally categorized into the following two dimensions:
\begin{itemize}
    \item \textbf{Model Capability}: We assess the model's performance on human exam questions, its proficiency in mathematics, its ability to follow multi-turn instructions, and its translation accuracy of different language pairs.
    \item \textbf{Model Trustworthiness}: We evaluate the model's safety and tendency to hallucinate, particularly in the context of Southeast Asia.
\end{itemize}

\subsection{Model Capability}
\subsubsection{Multilingual World Knowledge}

\paragraph{Dataset} We utilized the M3Exam dataset \cite{m3exam}, comprising real human exam questions collected from different countries and spanning different subjects and educational stages. This dataset effectively tests the model's multilingual world knowledge in a manner more akin to real-world settings. We take the questions in English (en), Chinese (zh), Indonesian (id), Vietnamese (vi), and Thai (th). We also employ the translated MMLU \cite{mmlu} questions for evaluation, which primarily tests the cross-lingual alignment of the model as the required knowledge is still mainly Western-focused. For evaluation, we employ the SeaExam toolkit\footnote{\url{https://github.com/DAMO-NLP-SG/SeaExam}} and measured performance using accuracy as the metric.

\paragraph{Results} As shown in Table \ref{tab:world-knowledge} for results on M3Exam dataset, our models, SeaLLMs-v3-7B and SeaLLMs-v3-7B-Chat, demonstrate competitive performance, with SeaLLMs-v3-7B-Chat achieving the highest average score (0.692) and the highest average score for SEA languages (0.592). 
Compared to the previous version, SeaLLMs-7B-v2.5, our latest models show significant improvement in overall performance specifically in handling Southeast Asian languages. Furthermore, while the Qwen2-7B-Instruct model performs exceptionally well in English and Chinese, our models exhibit superior performance across a broader range of Southeast Asian languages, highlighting their enhanced multilingual capabilities.

Table \ref{tab:mmlu} shows the results of different models on the translated MMLU dataset, we can see that our SeaLLMs-v3-7B and SeaLLMs-v3-7B-Chat models also outperform other models, particularly in Southeast Asian languages. 
Compared to the previous SeaLLMs-7B-v2.5 version, our latest models show substantial improvements, particularly in handling Southeast Asian languages (with avg\_sea from 52.4 to 58.2).

\begin{table*}[h]
    \centering
    \resizebox{0.72\textwidth}{!}{
    \begin{tabular}{lccccccc}
        \toprule
        \textbf{MGSM} & \textbf{en} & \textbf{id} & \textbf{ms} & \textbf{th} & \textbf{vi} & \textbf{zh} & \textbf{avg} \\ 
        \midrule
        \textit{Few-shot setting} \\
        Gemma-7B & 64.8 & 41.2 & 43.2 & 38.0 & 34.0 & 39.6 & 43.5   \\
        Sailor-7B & 34.4 & 25.2 & 22.8 & 24.8 & 22.4 & 26.4 & 26.0   \\ 
        Meta-Llama-3-8B & 56.8 & 36.0 & 33.6 & 34.8 & 33.6 & 43.6 & 39.7 \\
        GLM-4-9B & 78.0 & 53.6 & \textbf{57.2} & 46.0 & \textbf{56.8} & 69.6 & 60.2 \\
        Qwen2-7B & \textbf{79.6} & 58.8 & 56.8 & 54.8 & 54.8 & 69.2 & 62.3   \\ 
        \textbf{SeaLLMs-v3-7B} & 78.8 & \textbf{59.2}  & 56.8 & \textbf{56.8} & 54.8 & \textbf{72.0} & \textbf{63.1} \\ \midrule
        \textit{Zero-shot setting} \\
        Gemma-1.1-7B-it & 58.8 & 32.4 & 34.8 & 31.2 & 39.6 & 35.2 & 38.7 \\ 
        Sailor-7B-Chat & 33.6 & 22.4 & 22.4 & 21.6 & 25.2 & 29.2 & 25.7 \\ 
        SeaLLM-7B-v2.5 & 79.6 & 69.2 & \textbf{70.8} & 61.2 & 66.8 & 62.4 & 68.3 \\ 
        Meta-Llama-3-8B-Instruct & 77.6 & 48.0 & 57.6 & 56.0 & 46.8 & 58.8 & 57.5 \\ 
        GLM-4-9B-Chat & 72.8 & 53.6 & 53.6 & 34.8 & 52.4 & 70.8 & 56.3 \\ 
        Qwen2-7B-Instruct &\textbf{ 82.0} & 66.4 & 62.4 & 58.4 & 64.4 & 76.8 & 68.4 \\ 
        \textbf{SeaLLMs-v3-7B-Chat} & 74.8 & \textbf{71.2} & \textbf{70.8} &\textbf{ 71.2} & \textbf{71.2} & \textbf{79.6} & \textbf{73.1} \\ 
        \bottomrule
    \end{tabular}
    }
    \caption{Results of multilingual math with the MGSM benchmark.}
    \label{tab:math}
\end{table*}

\begin{table*}[t]
    \centering
    \resizebox{0.95\textwidth}{!}{
    \begin{tabular}{lcccccccccc}
    \toprule
    \multirow{2}{*}{\textbf{\normalsize{Model}}} & \multicolumn{3}{c}{\textbf{\normalsize{id}}} & \multicolumn{3}{c}{\textbf{\normalsize{th}}} & \multicolumn{3}{c}{\textbf{\normalsize{vi}}} & \multirow{2}{*}{\textbf{\normalsize{avg}}} \\
    \cmidrule(lr){2-4} \cmidrule(lr){5-7} \cmidrule(lr){8-10}
    & \textbf{turn1} & \textbf{turn2} & \textbf{avg} & \textbf{turn1} & \textbf{turn2} & \textbf{avg} & \textbf{turn1} & \textbf{turn2} & \textbf{avg} & \\
    \midrule
        Sailor-7B-Chat & 4.60 & 4.04 & 4.32 & 3.94 & 3.17 & 3.56 & 4.82 & 3.62 & 4.22 & 4.03 \\ 
        SeaLLM-7B-v2.5 & 6.27 & 4.96 & 5.62 & 5.79 & 3.82 & 4.81 & 6.02 & 4.02 & 5.02 & 5.15 \\ 
        Sailor-14B-Chat & 5.26 & 5.53 & 5.40 & 4.62 & 4.36 & 4.49 & 5.31 & 4.74 & 5.03 & 4.97 \\ 
        Qwen2-7B-Instruct & 5.93 & 5.84 & 5.89 & 5.47 & 5.20 & 5.34 & 6.17 & 5.60 & 5.89 & 5.70 \\ 
        \textbf{SeaLLMs-v3-7B-Chat} & \textbf{6.73} & \textbf{6.59} & \textbf{6.66} & \textbf{6.48} & \textbf{5.90} & \textbf{6.19} & \textbf{6.34} & \textbf{5.79} & \textbf{6.07} & \textbf{6.31} \\ 
    \bottomrule
    \end{tabular}
    }
    \caption{Results of multilingual instruction-following with SeaBench Benchmark}
    \label{tab:seabench}
\end{table*}

\begin{table*}[h]
    \centering
    \resizebox{\textwidth}{!}{%
    \begin{tabular}{lccccccccccccc}
        \toprule
      \textbf{\normalsize{Model}} & \textbf{\normalsize{en}} & \textbf{\normalsize{id}} & \textbf{\normalsize{jv}} & \textbf{\normalsize{km}} & \textbf{\normalsize{lo}} & \textbf{\normalsize{ms}} & \textbf{\normalsize{my}} & \textbf{\normalsize{ta}} & \textbf{\normalsize{th}} & \textbf{\normalsize{tl}} & \textbf{\normalsize{vi}} & \textbf{\normalsize{zh}} & \textbf{\normalsize{avg}} \\ 
        \midrule
        Sailor-7B-Chat & 49.40 & 49.78 & 28.33 & 2.68 & 6.85 & 47.75 & 5.35 & 18.23 & 38.92 & 29.00 & 41.76 & 20.87 & 28.24 \\
        SeaLLM-7B-v2.5 & \textbf{55.09} & \textbf{53.71} & 18.13 & 18.09 & 15.53 & \textbf{51.33} & 19.71 & 26.10 & 40.55 & \textbf{45.58} & \textbf{44.56} & 24.18 & 34.38 \\
        Meta-Llama-3-8B-Instruct & 51.54 & 49.03 & 22.46 & 15.34 & 5.42 & 46.72 & 21.24 & \textbf{32.09} & 35.75 & 40.80 & 39.31 & 14.87 & 31.22 \\
        Qwen2-7B-Instruct & 50.36 & 47.55 & 29.36 & 19.26 & 11.06 & 42.43 & 19.33 & 20.04 & 36.07 & 37.91 & 39.63 & 22.87 & 31.32 \\
        \textbf{SeaLLMs-v3-7B-Chat} & 54.68 & 52.52 & \textbf{29.86} & \textbf{27.30} & \textbf{26.34} & 45.04 & \textbf{21.54} & 31.93 & \textbf{41.52} & 38.51 & 43.78 & \textbf{26.10} & \textbf{36.52} \\
        \bottomrule
    \end{tabular}
    }
    \caption{Results of translation with Flores-200.}
    \label{tab:translation}
\end{table*}

\subsubsection{Multilingual Math}
\paragraph{Dataset} We assess the multilingual mathematics capabilities using the MGSM dataset \cite{mgsm-dataset}. Originally, MGSM comprises testing samples solely in English, Chinese and Thai. To extend this dataset to other SEA languages, specifically Indonesian, Malay, and Vietnamese, we utilize Google Translate to translate the original English questions to those questions. It is important to note that in our translations, we adhere to the numerical notation conventions of each respective country. For instance, in both Indonesian and Vietnamese, dots are used as thousands separators, and commas as decimal separators, which is the reverse of the English numeral system. We follow the same convention when evaluating the model's generations.

\paragraph{Results} Table \ref{tab:math} presents the evaluation results on the MGSM benchmark, under both the few-shot setting (for testing base versions) and the zero-shot setting (for testing chat versions).
In the few-shot setting, SeaLLMs-v3-7B demonstrates the highest average score (63.1), outperforming other models such as Qwen2-7B (62.3) and GLM-4-9b (60.2), particularly excelling in Indonesian and Thai. In the zero-shot setting, the chat version of SeaLLMs-v3-7B-Chat achieves the highest average score (73.1), showing strong performance across all languages. This highlights SeaLLMs-v3-7B-Chat's superior adaptability and robustness in multilingual math tasks compared to its counterparts like Qwen2-7B-instruct (68.4).

\subsubsection{Multilingual Instruction-following}
\paragraph{Dataset}
As there is no publicly available dataset for testing the model's multi-turn instruction-following capability in SEA languages, we construct our own benchmark, namely SeaBench\footnote{It will be publicly available at \url{https://huggingface.co/datasets/SeaLLMs/SeaBench}}, for such evaluation.
SeaBench consists of multi-turn human instructions spanning various task types for Indonesian, Vietnamese, and Thai. Following MT-Bench \cite{mtbench}, we consider 8 task types including writing, roleplay, extraction, reasoning, math, coding, knowledge I (STEM), and knowledge II (humanities/social science). Additionally, considering the characteristics of the multilingual setting, we include two more task types: safety and life. The safety task tests whether the model will respond to unsafe queries in a local context, while the life task includes questions likely asked in real-life settings, which might be informally written or even ambiguous.

All questions are manually written by native speakers of each language. During construction, we instructed the annotators to ensure the questions were as localized as possible, e.g., using local entities, concepts, and knowledge in the questions. Furthermore, reference answers have been constructed to ensure fair judgment.

\paragraph{Evaluation Details} Given the two-turn questions, the model under testing generates two-turn responses in a multi-turn format. These responses are then graded by a stronger LLM (GPT-4o was used in our experiments) using the reference answer to the original questions. The scores are then assigned to each turn of the response.

\paragraph{Results}
As shown in Table \ref{tab:seabench}, SeaLLMs-v3-7B-Chat outperforms all other models in multilingual instruction-following across Indonesian (id), Thai (th), and Vietnamese (vi). It achieves the highest average scores in both individual turns and overall averages for each language. Specifically, SeaLLMs-v3-7B-Chat surpasses the previous version, SeaLLMs-7B-v2.5, by a significant margin (6.31 vs 5.15) and outperforms the strongest baseline model Qwen2-7B-Instruct (6.31 vs 5.70). These results highlight SeaLLMs-v3-7B-Chat's superior ability to generate more coherent and contextually appropriate multi-turn responses.

\begin{table*}[t]
    \centering
    \resizebox{0.8\textwidth}{!}{
    \begin{tabular}{lccccccc}
    \toprule
    \textbf{Model} & \textbf{en} & \textbf{zh} & \textbf{vi} & \textbf{th} & \textbf{id} & \textbf{avg} & \textbf{avg\_sea} \\
    \midrule
    Gemma-1.1-7B-it & 53.61 & 28.22 & 26.18 & 21.28 & 30.39 & 31.94 &    25.95 \\
     Sailor-7B-Chat & 33.76 & 18.82 & 5.19 &  9.68 & 16.42 & 16.78 &    10.43 \\
     SeaLLM-7B-v2.5 & 13.10 &  1.53 &  3.24 & 19.58 &  0.78 &  7.65 &     7.87 \\
    Meta-Llama-3-8B-Instruct & \textbf{72.23} &  0.00 &  1.23 &  0.80 &  3.91 & 15.63 &     1.98 \\
      GLM-4-9B-Chat  & 45.02 & 40.98 & 21.48 &  5.42 &  2.37 & 23.05 &     9.76 \\
  Qwen2-7B-Instruct & 63.74 & 35.75 & 52.86 & 46.42 & 55.93 & 50.94 &    51.74 \\
    \midrule
 \textbf{SeaLLMs-v3-7B-Chat} & 71.13 & \textbf{77.17} & \textbf{78.18} & \textbf{61.64} & \textbf{67.61} & \textbf{71.14} &    \textbf{69.14} \\
    \bottomrule
    \end{tabular}
    }
    \caption{Performance in refusing questions about non-existing entities on SeaRefuse-G.}
    \label{tab:hallucination-M}
\end{table*}

\begin{table*}[t]
    \centering
    \resizebox{0.8\textwidth}{!}{
    \begin{tabular}{lccccccc}
    \toprule
    \textbf{Model} & \textbf{en} & \textbf{zh} & \textbf{vi} & \textbf{th} & \textbf{id} & \textbf{avg} & \textbf{avg\_sea} \\
    \midrule
    Gemma-1.1-7B-it & 51.95 & 29.92 & 12.07 & 30.16 & 35.48 & 31.92 &    25.90 \\
Sailor-7B-Chat & 33.87 & 15.79 & 11.32 & 12.96 & 31.67 & 21.12 & 18.65 \\
SeaLLM-7B-v2.5 & 10.91 & 3.92 & 11.32 & 49.66 & 31.15 & 21.39 & 30.71 \\
Meta-Llama-3-8B-Instruct & \textbf{73.17} & 0.00 & 0.00 & 0.00 & 11.21 & 16.88 & 3.74 \\
GLM-4-9B-Chat & 35.38 & 52.50 & 20.17 & 5.77 & 9.52 & 24.67 & 11.82 \\
Qwen2-7B-Instruct & 58.50 & 42.75 & 62.82 & 60.53 & 63.51 & 57.62 & 62.29 \\
    \midrule
 \textbf{SeaLLMs-v3-7B-Chat} & 68.54 & \textbf{81.82} & \textbf{83.84} & \textbf{84.58} & \textbf{89.66} & \textbf{81.69} &    \textbf{86.03} \\
    \bottomrule
    \end{tabular}
    }
    \caption{Performance in refusing questions about non-existing entities on SeaRefuse-H.}
    \label{tab:hallucination-H}
\end{table*}

\subsubsection{Translation}
\paragraph{Dataset}
We evaluate the machine translation performances with the test set of Flores-200 \cite{flores-dataset}. We choose all 12 languages for a comprehensive evaluation, including Burmese (my), Chinese (zh), English (en), Indonesian (id), Javanese (jv), Khmer (km), Lao (lo), Malay (ms), Tagalog (tl), Tamil (ta), Thai (th), and Vietnamese (vi). We translate between each pair of languages and report the average 0-shot  chrF scores after averaging the results for target languages. 

\paragraph{Results}
As shown in Table \ref{tab:translation}, SeaLLMs-v3-7B-Chat outperforms other models in machine translation, achieving an average chrF score of 36.52. It excels particularly in Javanese, Khmer, Lao, Burmese, Thai, and Chinese, consistently achieving the highest scores in these languages. Compared to its predecessor, SeaLLMs-7B-v2.5, which has an average score of 34.38, SeaLLMs-v3-7B-Chat shows clear improvement. Additionally, SeaLLMs-v3-7B-Chat surpasses strong baselines like Meta-Llama-3-8B-Instruct and Qwen2-7B-Instruct, with average scores of 31.22 and 31.32, respectively. Notably, the model's performance in translating low-resource languages, such as Khmer (27.30) and Lao (26.34), highlights its robustness and effectiveness in handling diverse and challenging translation tasks. This consistent performance across multiple languages underscores the model's versatility and capability in low-resource language translation settings.

\subsection{Model Trustworthiness}
\subsubsection{Hallucination}
A trustworthy LLM should only answer the questions that it knows and abstain from answering questions that it does not know. Previous studies reveal that recent LLMs are prone to answering questions that exceed their knowledge boundaries, leading to hallucinated responses~\cite{DBLP:journals/corr/alignment-for-honesty-2023,zhang-etal-2024-r}. However, evaluating a model's ability to refuse questions it doesn't know is challenging. This requires distinguishing the model's knowledge boundaries, which is difficult since most existing LLMs do not provide transparency in their pre-training data. 

To address this challenge, we propose a novel \textbf{SeaRefuse} evaluation benchmark, which consists of unanswerable factoid questions about non-existing entities and answerable factoid questions in SEA languages\footnote{It will be publicly available at \url{https://huggingface.co/datasets/SeaLLMs/SeaRefuse}}. Unanswerable questions about non-existent entities are designed to surpass the knowledge boundaries of LLMs. 
Our benchmark includes two test sets: \textbf{SeaRefuse-G} and \textbf{SeaRefuse-H}. In the SeaRefuse-G test set, the unanswerable questions are generated by prompting GPT-4o. In the SeaRefuse-H test set, our linguists annotate the unanswerable questions by refining machine-generated unanswerable questions. The answerable questions in both SeaRefuse-G and SeaRefuse-H are collected from existing factoid QA datasets including ParaRel~\cite{zhang-etal-2024-r,DBLP:journals/tacl/ElazarKRRHSG21}, NLPCC-KBQA~\cite{NLPCC-KBQ16,NLPCC-KBQ17}, and TyDi QA~\cite{tydiqa}. For the SeaRefuse-G tese set, each language has 500 answerable and 500 unanswerable questions, except for Vietnamese, which has 483 of each. In the SeaRefuse-H dataset, each language contains 100 answerable and 100 unanswerable questions.

\begin{table}[ht]
\small
\centering
\begin{tabular}{@{}c | c | c@{}}
\toprule
& \begin{tabular}[c]{@{}c@{}}\textbf{Answerable}\\ \textbf{question}\end{tabular} & \begin{tabular}[c]{@{}c@{}}\textbf{Unanswerable}\\ \textbf{question}\end{tabular} \\
\midrule
\textbf{Refused} & True-Positive & False-Positive \\\midrule
\textbf{Answered} & False-Negative & True-Negative \\
\bottomrule
\end{tabular}
\caption{The confusion matrix for the evaluation of the refusal ability of LLMs.}
\label{tab:confusion-matrix}
\end{table}

In evaluation, we report the F1-score of each model on correctly refusing questions about non-existing entities. We follow the confusion matrix depicted in Table~\ref{tab:confusion-matrix} to compute the F1 scores. 
We adopt a keyword-matching approach to determine whether a model refuses to answer a factoid question. Specifically, we work with professional and native linguists to devise a set of refusal keywords for English, Chinese, Vietnamese, Indonesian, and Thai, respectively. If the generated response contains any of these refusal keywords, we assume that the response is a refusal response. 

The experiment results on SeaRefuse-G and SeaRefuse-H are shown in Table~\ref{tab:hallucination-M} and Table~\ref{tab:hallucination-H}, respectively. We observe that SeaLLMs-v3-7B-Chat outperforms all other baseline models by a large margin in zh, vi, th, and id languages. 
In English, the performance of SeaLLMs-v3-7B-Chat is competitive with Llama-3-8B-Instruct. These results demonstrate the capability of SeaLLMs-v3 to refuse questions that it does not know.

\subsubsection{Safety}
\begin{table}[th]
    \centering
    \resizebox{\columnwidth}{!}{
    \begin{tabular}{l|ccccc|c}
        \toprule
        \textbf{Model} & \textbf{en} & \textbf{jv} & \textbf{th} & \textbf{vi} & \textbf{zh} & \textbf{avg} \\
        \midrule
    Sailor-7B-Chat & 78.7 & 54.9 & 62.2 & 67.6 & 76.2 & 67.9 \\
    Meta-Llama-3-8B-Instruct & 88.3 & 26.4 & 71.1 & 69.8 & 77.1 & 66.5 \\
    Sailor-14B-Chat & 86.9 & 30.5 & 53.7 & 60.9 & 72.7 & 60.9 \\
    GLM-4-9B-Chat & 77.1 & 21.3 & 30.2 & 60.6 & 74.9 & 52.8 \\
    Qwen2-7B-Instruct & 88.6 & 43.8 & 63.8 & 73.0 & 87.3 & 71.3 \\
    \midrule
    \textbf{SeaLLMs-v3-7B-Chat} & \textbf{88.9} & \textbf{60.0} & \textbf{73.3} & \textbf{83.8} & \textbf{92.7} & \textbf{79.7} \\
    \bottomrule
    \end{tabular}
    }
    \caption{Safety performance of different models}
    \label{tab:safety_performance}
\end{table}

To evaluate the models' safety capabilities, we use the questions of SEA languages from the MultiJail dataset \cite{multilingual-jailbreak}, which includes English (en), Javanese (jv), Thai (th), Vietnamese (vi), and Chinese (zh). Each question in the dataset is potentially malicious, and the model should refuse to answer them. To determine whether the model's response is safe, we first translate the response into English and then prompt GPT-4o to check if the translated response is harmful. The results are reported as the safe rate of the responses.

Table \ref{tab:safety_performance} presents the safety capabilities of various models evaluated with the MultiJail dataset. Notably, SeaLLMs-v3-7B-Chat outperforms all other models with an average safe rate of 79.7\%, demonstrating robust performance across all languages, particularly excelling in Vietnamese (83.8\%) and Chinese (92.7\%). In comparison, Qwen2-7B-Instruct follows with a distant second average of 71.3\%, with its highest safe rate in Chinese (87.3\%). Other models like Sailor-7B-Chat and Llama-3-8B-Instruct also show competitive performance but lag behind in consistency across languages. 
Notably, the exceptional performance of SeaLLMs-v3 in the three Southeast Asian languages (jv, th, and vi) underscores SeaLLM's effective design, which caters to the linguistic nuances of this region.

\section{Conclusion}
SeaLLMs 3 represents a significant advancement in the development of large language models for Southeast Asian languages, addressing the region's unique linguistic and cultural challenges. By adopting an efficient language enhancement approach and constructing a comprehensive instruction tuning dataset, SeaLLMs 3 achieves state-of-the-art performance while maintaining cost-effectiveness. Our commitment to reliability and safety, providing contextually appropriate responses, further strengthens the model's applicability and trustworthiness. The open-sourcing of both foundational and chat models ensures that SeaLLMs 3 is accessible for a wide range of applications, fostering further innovation and inclusivity in AI development for Southeast Asia.

\section*{Acknowledgments}

We would like to express our special thanks to our professional and native linguists, Tantong Champaiboon, Nguyen Ngoc Yen Nhi and Tara Devina Putri, who helped build, evaluate, and fact-check our sampled pretraining and SFT dataset as well as evaluating our models across different aspects, especially safety.

\bibliography{custom}

\begin{thebibliography}{41}
\providecommand{\natexlab}[1]{#1}

\bibitem[{Ahuja et~al.(2023)Ahuja, Diddee, Hada, Ochieng, Ramesh, Jain, Nambi, Ganu, Segal, Ahmed, Bali, and Sitaram}]{multilingual-eval-emnlp23}
Kabir Ahuja, Harshita Diddee, Rishav Hada, Millicent Ochieng, Krithika Ramesh, Prachi Jain, Akshay~Uttama Nambi, Tanuja Ganu, Sameer Segal, Mohamed Ahmed, Kalika Bali, and Sunayana Sitaram. 2023.
\newblock \href {https://doi.org/10.18653/v1/2023.emnlp-main.258} {{MEGA:} multilingual evaluation of generative {AI}}.
\newblock In \emph{Proceedings of the 2023 Conference on Empirical Methods in Natural Language Processing, {EMNLP} 2023}, pages 4232--4267.

\bibitem[{{AI Singapore}(2023)}]{sea_lion_2023}
{AI Singapore}. 2023.
\newblock Sea-lion (southeast asian languages in one network): A family of large language models for southeast asia.
\newblock \url{https://github.com/aisingapore/sealion}.

\bibitem[{Aji et~al.(2022)Aji, Winata, Koto, Cahyawijaya, Romadhony, Mahendra, Kurniawan, Moeljadi, Prasojo, Baldwin, Lau, and Ruder}]{acl22-indonlp}
Alham~Fikri Aji, Genta~Indra Winata, Fajri Koto, Samuel Cahyawijaya, Ade Romadhony, Rahmad Mahendra, Kemal Kurniawan, David Moeljadi, Radityo~Eko Prasojo, Timothy Baldwin, Jey~Han Lau, and Sebastian Ruder. 2022.
\newblock \href {https://doi.org/10.18653/v1/2022.acl-long.500} {One country, 700+ languages: {NLP} challenges for underrepresented languages and dialects in indonesia}.
\newblock In \emph{Proceedings of the 60th Annual Meeting of the Association for Computational Linguistics (Volume 1: Long Papers), {ACL} 2022}, pages 7226--7249. Association for Computational Linguistics.

\bibitem[{Anil et~al.(2023)Anil, Borgeaud, Wu, Alayrac, Yu, Soricut, Schalkwyk, Dai, Hauth, Millican, Silver, Petrov, Johnson, Antonoglou, Schrittwieser, Glaese, Chen, Pitler, Lillicrap, Lazaridou, Firat, Molloy, Isard, Barham, Hennigan, Lee, Viola, Reynolds, Xu, Doherty, Collins, Meyer, Rutherford, Moreira, Ayoub, Goel, Tucker, Piqueras, Krikun, Barr, Savinov, Danihelka, Roelofs, White, Andreassen, von Glehn, Yagati, Kazemi, Gonzalez, Khalman, Sygnowski, and et~al.}]{gemini}
Rohan Anil, Sebastian Borgeaud, Yonghui Wu, Jean{-}Baptiste Alayrac, Jiahui Yu, Radu Soricut, Johan Schalkwyk, Andrew~M. Dai, Anja Hauth, Katie Millican, David Silver, Slav Petrov, Melvin Johnson, Ioannis Antonoglou, Julian Schrittwieser, Amelia Glaese, Jilin Chen, Emily Pitler, Timothy~P. Lillicrap, Angeliki Lazaridou, Orhan Firat, James Molloy, Michael Isard, Paul~Ronald Barham, Tom Hennigan, Benjamin Lee, Fabio Viola, Malcolm Reynolds, Yuanzhong Xu, Ryan Doherty, Eli Collins, Clemens Meyer, Eliza Rutherford, Erica Moreira, Kareem Ayoub, Megha Goel, George Tucker, Enrique Piqueras, Maxim Krikun, Iain Barr, Nikolay Savinov, Ivo Danihelka, Becca Roelofs, Ana{\"{\i}}s White, Anders Andreassen, Tamara von Glehn, Lakshman Yagati, Mehran Kazemi, Lucas Gonzalez, Misha Khalman, Jakub Sygnowski, and et~al. 2023.
\newblock \href {https://doi.org/10.48550/arXiv.2312.11805} {Gemini: {A} family of highly capable multimodal models}.
\newblock \emph{CoRR}, abs/2312.11805.

\bibitem[{Aryabumi et~al.(2024)Aryabumi, Dang, Talupuru, Dash, Cairuz, Lin, Venkitesh, Smith, Campos, Tan, Marchisio, Bartolo, Ruder, Locatelli, Kreutzer, Frosst, Gomez, Blunsom, Fadaee, {\"{U}}st{\"{u}}n, and Hooker}]{aya-model}
Viraat Aryabumi, John Dang, Dwarak Talupuru, Saurabh Dash, David Cairuz, Hangyu Lin, Bharat Venkitesh, Madeline Smith, Jon~Ander Campos, Yi~Chern Tan, Kelly Marchisio, Max Bartolo, Sebastian Ruder, Acyr Locatelli, Julia Kreutzer, Nick Frosst, Aidan~N. Gomez, Phil Blunsom, Marzieh Fadaee, Ahmet {\"{U}}st{\"{u}}n, and Sara Hooker. 2024.
\newblock \href {https://doi.org/10.48550/arXiv.2405.15032} {Aya 23: Open weight releases to further multilingual progress}.
\newblock \emph{CoRR}, abs/2405.15032.

\bibitem[{Bai et~al.(2023)Bai, Bai, Chu, Cui, Dang, Deng, Fan, Ge, Han, Huang, Hui, Ji, Li, Lin, Lin, Liu, Liu, Lu, Lu, Ma, Men, Ren, Ren, Tan, Tan, Tu, Wang, Wang, Wang, Wu, Xu, Xu, Yang, Yang, Yang, Yang, Yao, Yu, Yuan, Yuan, Zhang, Zhang, Zhang, Zhang, Zhou, Zhou, Zhou, and Zhu}]{qwen-1}
Jinze Bai, Shuai Bai, Yunfei Chu, Zeyu Cui, Kai Dang, Xiaodong Deng, Yang Fan, Wenbin Ge, Yu~Han, Fei Huang, Binyuan Hui, Luo Ji, Mei Li, Junyang Lin, Runji Lin, Dayiheng Liu, Gao Liu, Chengqiang Lu, Keming Lu, Jianxin Ma, Rui Men, Xingzhang Ren, Xuancheng Ren, Chuanqi Tan, Sinan Tan, Jianhong Tu, Peng Wang, Shijie Wang, Wei Wang, Shengguang Wu, Benfeng Xu, Jin Xu, An~Yang, Hao Yang, Jian Yang, Shusheng Yang, Yang Yao, Bowen Yu, Hongyi Yuan, Zheng Yuan, Jianwei Zhang, Xingxuan Zhang, Yichang Zhang, Zhenru Zhang, Chang Zhou, Jingren Zhou, Xiaohuan Zhou, and Tianhang Zhu. 2023.
\newblock \href {https://doi.org/10.48550/arXiv.2309.16609} {Qwen technical report}.
\newblock \emph{CoRR}, abs/2309.16609.

\bibitem[{Ben~Allal et~al.(2024)Ben~Allal, Lozhkov, Penedo, Wolf, and von Werra}]{benallal2024cosmopedia}
Loubna Ben~Allal, Anton Lozhkov, Guilherme Penedo, Thomas Wolf, and Leandro von Werra. 2024.
\newblock \href {https://huggingface.co/datasets/HuggingFaceTB/cosmopedia} {Cosmopedia}.

\bibitem[{Clark et~al.(2020)Clark, Choi, Collins, Garrette, Kwiatkowski, Nikolaev, and Palomaki}]{tydiqa}
Jonathan~H. Clark, Eunsol Choi, Michael Collins, Dan Garrette, Tom Kwiatkowski, Vitaly Nikolaev, and Jennimaria Palomaki. 2020.
\newblock Tydi qa: A benchmark for information-seeking question answering in typologically diverse languages.
\newblock \emph{Transactions of the Association for Computational Linguistics}.

\bibitem[{Costa{-}juss{\`{a}} et~al.(2022)Costa{-}juss{\`{a}}, Cross, {\c{C}}elebi, Elbayad, Heafield, Heffernan, Kalbassi, Lam, Licht, Maillard, Sun, Wang, Wenzek, Youngblood, Akula, Barrault, Gonzalez, Hansanti, Hoffman, Jarrett, Sadagopan, Rowe, Spruit, Tran, Andrews, Ayan, Bhosale, Edunov, Fan, Gao, Goswami, Guzm{\'{a}}n, Koehn, Mourachko, Ropers, Saleem, Schwenk, and Wang}]{flores-dataset}
Marta~R. Costa{-}juss{\`{a}}, James Cross, Onur {\c{C}}elebi, Maha Elbayad, Kenneth Heafield, Kevin Heffernan, Elahe Kalbassi, Janice Lam, Daniel Licht, Jean Maillard, Anna~Y. Sun, Skyler Wang, Guillaume Wenzek, Al~Youngblood, Bapi Akula, Lo{\"{\i}}c Barrault, Gabriel~Mejia Gonzalez, Prangthip Hansanti, John Hoffman, Semarley Jarrett, Kaushik~Ram Sadagopan, Dirk Rowe, Shannon Spruit, Chau Tran, Pierre Andrews, Necip~Fazil Ayan, Shruti Bhosale, Sergey Edunov, Angela Fan, Cynthia Gao, Vedanuj Goswami, Francisco Guzm{\'{a}}n, Philipp Koehn, Alexandre Mourachko, Christophe Ropers, Safiyyah Saleem, Holger Schwenk, and Jeff Wang. 2022.
\newblock \href {https://doi.org/10.48550/arXiv.2207.04672} {No language left behind: Scaling human-centered machine translation}.
\newblock \emph{CoRR}, abs/2207.04672.

\bibitem[{Crawl()}]{ccnews}
Common Crawl.
\newblock \href {https://commoncrawl.org/overview} {Common crawl news}.

\bibitem[{Deng et~al.(2023)Deng, Zhang, Pan, and Bing}]{multilingual-jailbreak}
Yue Deng, Wenxuan Zhang, Sinno~Jialin Pan, and Lidong Bing. 2023.
\newblock \href {https://doi.org/10.48550/arXiv.2310.06474} {Multilingual jailbreak challenges in large language models}.
\newblock \emph{CoRR}, abs/2310.06474.

\bibitem[{Dou et~al.(2024)Dou, Liu, Zeng, Guo, Zhou, Lu, and Lin}]{sailor}
Longxu Dou, Qian Liu, Guangtao Zeng, Jia Guo, Jiahui Zhou, Wei Lu, and Min Lin. 2024.
\newblock \href {https://doi.org/10.48550/arXiv.2404.03608} {Sailor: Open language models for south-east asia}.
\newblock \emph{CoRR}, abs/2404.03608.

\bibitem[{Duan(2016)}]{NLPCC-KBQ16}
Nan Duan. 2016.
\newblock Overview of the nlpcc-iccpol 2016 shared task: Open domain chinese question answering.
\newblock In \emph{Natural Language Understanding and Intelligent Applications}, pages 942--948. Springer International Publishing.

\bibitem[{Duan and Tang(2018)}]{NLPCC-KBQ17}
Nan Duan and Duyu Tang. 2018.
\newblock Overview of the nlpcc 2017 shared task: Open domain chinese question answering.
\newblock In \emph{Natural Language Processing and Chinese Computing}, pages 954--961. Springer International Publishing.

\bibitem[{Elazar et~al.(2021)Elazar, Kassner, Ravfogel, Ravichander, Hovy, Sch{\"{u}}tze, and Goldberg}]{DBLP:journals/tacl/ElazarKRRHSG21}
Yanai Elazar, Nora Kassner, Shauli Ravfogel, Abhilasha Ravichander, Eduard~H. Hovy, Hinrich Sch{\"{u}}tze, and Yoav Goldberg. 2021.
\newblock \href {https://doi.org/10.1162/TACL\_A\_00410} {Measuring and improving consistency in pretrained language models}.
\newblock \emph{Trans. Assoc. Comput. Linguistics}, 9:1012--1031.

\bibitem[{Foundation()}]{wikidump}
Wikimedia Foundation.
\newblock \href {https://dumps.wikimedia.org} {Wikimedia downloads}.

\bibitem[{Hendrycks et~al.(2021)Hendrycks, Burns, Basart, Zou, Mazeika, Song, and Steinhardt}]{mmlu}
Dan Hendrycks, Collin Burns, Steven Basart, Andy Zou, Mantas Mazeika, Dawn Song, and Jacob Steinhardt. 2021.
\newblock \href {https://openreview.net/forum?id=d7KBjmI3GmQ} {Measuring massive multitask language understanding}.
\newblock In \emph{9th International Conference on Learning Representations, {ICLR} 2021}. OpenReview.net.

\bibitem[{Huang et~al.(2024)Huang, Mo, Li, Li, Zhang, Yi, Mao, Liu, Xu, Xu, Nie, and Liu}]{multilingual-survey-kaiyu}
Kaiyu Huang, Fengran Mo, Hongliang Li, You Li, Yuanchi Zhang, Weijian Yi, Yulong Mao, Jinchen Liu, Yuzhuang Xu, Jinan Xu, Jian{-}Yun Nie, and Yang Liu. 2024.
\newblock \href {https://doi.org/10.48550/arXiv.2405.10936} {A survey on large language models with multilingualism: Recent advances and new frontiers}.
\newblock \emph{CoRR}, abs/2405.10936.

\bibitem[{Jiang et~al.(2023)Jiang, Sablayrolles, Mensch, Bamford, Chaplot, de~Las~Casas, Bressand, Lengyel, Lample, Saulnier, Lavaud, Lachaux, Stock, Scao, Lavril, Wang, Lacroix, and Sayed}]{mistral-7b}
Albert~Q. Jiang, Alexandre Sablayrolles, Arthur Mensch, Chris Bamford, Devendra~Singh Chaplot, Diego de~Las~Casas, Florian Bressand, Gianna Lengyel, Guillaume Lample, Lucile Saulnier, L{\'{e}}lio~Renard Lavaud, Marie{-}Anne Lachaux, Pierre Stock, Teven~Le Scao, Thibaut Lavril, Thomas Wang, Timoth{\'{e}}e Lacroix, and William~El Sayed. 2023.
\newblock \href {https://doi.org/10.48550/arXiv.2310.06825} {Mistral 7b}.
\newblock \emph{CoRR}, abs/2310.06825.

\bibitem[{Kudugunta et~al.(2023)Kudugunta, Caswell, Zhang, Garcia, Choquette-Choo, Lee, Xin, Kusupati, Stella, Bapna, and Firat}]{kudugunta2023madlad400}
Sneha Kudugunta, Isaac Caswell, Biao Zhang, Xavier Garcia, Christopher~A. Choquette-Choo, Katherine Lee, Derrick Xin, Aditya Kusupati, Romi Stella, Ankur Bapna, and Orhan Firat. 2023.
\newblock \href {https://arxiv.org/abs/2309.04662} {Madlad-400: A multilingual and document-level large audited dataset}.
\newblock \emph{Preprint}, arXiv:2309.04662.

\bibitem[{Liu et~al.(2024)Liu, Zhang, Zhao, Luu, and Bing}]{translation-analysis}
Chaoqun Liu, Wenxuan Zhang, Yiran Zhao, Anh~Tuan Luu, and Lidong Bing. 2024.
\newblock \href {https://doi.org/10.48550/arXiv.2403.10258} {Is translation all you need? {A} study on solving multilingual tasks with large language models}.
\newblock \emph{CoRR}, abs/2403.10258.

\bibitem[{Madaan et~al.(2023)Madaan, Tandon, Gupta, Hallinan, Gao, Wiegreffe, Alon, Dziri, Prabhumoye, Yang, Gupta, Majumder, Hermann, Welleck, Yazdanbakhsh, and Clark}]{nips23-refine}
Aman Madaan, Niket Tandon, Prakhar Gupta, Skyler Hallinan, Luyu Gao, Sarah Wiegreffe, Uri Alon, Nouha Dziri, Shrimai Prabhumoye, Yiming Yang, Shashank Gupta, Bodhisattwa~Prasad Majumder, Katherine Hermann, Sean Welleck, Amir Yazdanbakhsh, and Peter Clark. 2023.
\newblock \href {http://papers.nips.cc/paper\_files/paper/2023/hash/91edff07232fb1b55a505a9e9f6c0ff3-Abstract-Conference.html} {Self-refine: Iterative refinement with self-feedback}.
\newblock In \emph{Advances in Neural Information Processing Systems 36: Annual Conference on Neural Information Processing Systems 2023, NeurIPS 2023}.

\bibitem[{Mesnard et~al.(2024)Mesnard, Hardin, Dadashi, Bhupatiraju, Pathak, Sifre, Rivi{\`{e}}re, Kale, Love, Tafti, Hussenot, Chowdhery, Roberts, Barua, Botev, Castro{-}Ros, Slone, H{\'{e}}liou, Tacchetti, Bulanova, Paterson, Tsai, Shahriari, Lan, Choquette{-}Choo, Crepy, Cer, Ippolito, Reid, Buchatskaya, Ni, Noland, Yan, Tucker, Muraru, Rozhdestvenskiy, Michalewski, Tenney, Grishchenko, Austin, Keeling, Labanowski, Lespiau, Stanway, Brennan, Chen, Ferret, Chiu, and et~al.}]{gemma-1}
Thomas Mesnard, Cassidy Hardin, Robert Dadashi, Surya Bhupatiraju, Shreya Pathak, Laurent Sifre, Morgane Rivi{\`{e}}re, Mihir~Sanjay Kale, Juliette Love, Pouya Tafti, L{\'{e}}onard Hussenot, Aakanksha Chowdhery, Adam Roberts, Aditya Barua, Alex Botev, Alex Castro{-}Ros, Ambrose Slone, Am{\'{e}}lie H{\'{e}}liou, Andrea Tacchetti, Anna Bulanova, Antonia Paterson, Beth Tsai, Bobak Shahriari, Charline~Le Lan, Christopher~A. Choquette{-}Choo, Cl{\'{e}}ment Crepy, Daniel Cer, Daphne Ippolito, David Reid, Elena Buchatskaya, Eric Ni, Eric Noland, Geng Yan, George Tucker, George{-}Cristian Muraru, Grigory Rozhdestvenskiy, Henryk Michalewski, Ian Tenney, Ivan Grishchenko, Jacob Austin, James Keeling, Jane Labanowski, Jean{-}Baptiste Lespiau, Jeff Stanway, Jenny Brennan, Jeremy Chen, Johan Ferret, Justin Chiu, and et~al. 2024.
\newblock \href {https://doi.org/10.48550/arXiv.2403.08295} {Gemma: Open models based on gemini research and technology}.
\newblock \emph{CoRR}, abs/2403.08295.

\bibitem[{Nguyen et~al.(2023{\natexlab{a}})Nguyen, Van~Nguyen, Lai, Man, Ngo, Dernoncourt, Rossi, and Nguyen}]{nguyen2023culturax}
Thuat Nguyen, Chien Van~Nguyen, Viet~Dac Lai, Hieu Man, Nghia~Trung Ngo, Franck Dernoncourt, Ryan~A Rossi, and Thien~Huu Nguyen. 2023{\natexlab{a}}.
\newblock Culturax: A cleaned, enormous, and multilingual dataset for large language models in 167 languages.
\newblock \emph{arXiv preprint arXiv:2309.09400}.

\bibitem[{Nguyen et~al.(2023{\natexlab{b}})Nguyen, Aljunied, Joty, and Bing}]{acl24-phi}
Xuan{-}Phi Nguyen, Sharifah~Mahani Aljunied, Shafiq Joty, and Lidong Bing. 2023{\natexlab{b}}.
\newblock \href {https://doi.org/10.48550/arXiv.2306.11372} {Democratizing llms for low-resource languages by leveraging their english dominant abilities with linguistically-diverse prompts}.
\newblock \emph{CoRR}, abs/2306.11372.

\bibitem[{Nguyen et~al.(2023{\natexlab{c}})Nguyen, Zhang, Li, Aljunied, Tan, Cheng, Chen, Deng, Yang, Liu, Zhang, and Bing}]{seallms}
Xuan{-}Phi Nguyen, Wenxuan Zhang, Xin Li, Mahani Aljunied, Qingyu Tan, Liying Cheng, Guanzheng Chen, Yue Deng, Sen Yang, Chaoqun Liu, Hang Zhang, and Lidong Bing. 2023{\natexlab{c}}.
\newblock \href {https://doi.org/10.48550/arXiv.2312.00738} {Seallms - large language models for southeast asia}.
\newblock \emph{CoRR}, abs/2312.00738.

\bibitem[{OpenAI(2023)}]{gpt-4}
OpenAI. 2023.
\newblock \href {https://doi.org/10.48550/arXiv.2303.08774} {{GPT-4} technical report}.
\newblock \emph{CoRR}, abs/2303.08774.

\bibitem[{Qin et~al.(2024)Qin, Chen, Zhou, Chen, Li, Liao, Li, Che, and Yu}]{multilingual-survey-libo}
Libo Qin, Qiguang Chen, Yuhang Zhou, Zhi Chen, Yinghui Li, Lizi Liao, Min Li, Wanxiang Che, and Philip~S. Yu. 2024.
\newblock \href {https://doi.org/10.48550/arXiv.2404.04925} {Multilingual large language model: {A} survey of resources, taxonomy and frontiers}.
\newblock \emph{CoRR}, abs/2404.04925.

\bibitem[{Shi et~al.(2023)Shi, Suzgun, Freitag, Wang, Srivats, Vosoughi, Chung, Tay, Ruder, Zhou, Das, and Wei}]{mgsm-dataset}
Freda Shi, Mirac Suzgun, Markus Freitag, Xuezhi Wang, Suraj Srivats, Soroush Vosoughi, Hyung~Won Chung, Yi~Tay, Sebastian Ruder, Denny Zhou, Dipanjan Das, and Jason Wei. 2023.
\newblock \href {https://openreview.net/pdf?id=fR3wGCk-IXp} {Language models are multilingual chain-of-thought reasoners}.
\newblock In \emph{The Eleventh International Conference on Learning Representations, {ICLR} 2023}. OpenReview.net.

\bibitem[{Taori et~al.(2023)Taori, Gulrajani, Zhang, Dubois, Li, Guestrin, Liang, and Hashimoto}]{alpaca}
Rohan Taori, Ishaan Gulrajani, Tianyi Zhang, Yann Dubois, Xuechen Li, Carlos Guestrin, Percy Liang, and Tatsunori~B. Hashimoto. 2023.
\newblock \href {https://github.com/tatsu-lab/stanford_alpaca} {Stanford alpaca: An instruction-following llama model}.

\bibitem[{Touvron et~al.(2023{\natexlab{a}})Touvron, Lavril, Izacard, Martinet, Lachaux, Lacroix, Rozi{\`{e}}re, Goyal, Hambro, Azhar, Rodriguez, Joulin, Grave, and Lample}]{llama-1}
Hugo Touvron, Thibaut Lavril, Gautier Izacard, Xavier Martinet, Marie{-}Anne Lachaux, Timoth{\'{e}}e Lacroix, Baptiste Rozi{\`{e}}re, Naman Goyal, Eric Hambro, Faisal Azhar, Aur{\'{e}}lien Rodriguez, Armand Joulin, Edouard Grave, and Guillaume Lample. 2023{\natexlab{a}}.
\newblock \href {https://doi.org/10.48550/arXiv.2302.13971} {Llama: Open and efficient foundation language models}.
\newblock \emph{CoRR}, abs/2302.13971.

\bibitem[{Touvron et~al.(2023{\natexlab{b}})Touvron, Martin, Stone, Albert, Almahairi, Babaei, Bashlykov, Batra, Bhargava, Bhosale, Bikel, Blecher, Canton{-}Ferrer, Chen, Cucurull, Esiobu, Fernandes, Fu, Fu, Fuller, Gao, Goswami, Goyal, Hartshorn, Hosseini, Hou, Inan, Kardas, Kerkez, Khabsa, Kloumann, Korenev, Koura, Lachaux, Lavril, Lee, Liskovich, Lu, Mao, Martinet, Mihaylov, Mishra, Molybog, Nie, Poulton, Reizenstein, Rungta, Saladi, Schelten, Silva, Smith, Subramanian, Tan, Tang, Taylor, Williams, Kuan, Xu, Yan, Zarov, Zhang, Fan, Kambadur, Narang, Rodriguez, Stojnic, Edunov, and Scialom}]{llama-2}
Hugo Touvron, Louis Martin, Kevin Stone, Peter Albert, Amjad Almahairi, Yasmine Babaei, Nikolay Bashlykov, Soumya Batra, Prajjwal Bhargava, Shruti Bhosale, Dan Bikel, Lukas Blecher, Cristian Canton{-}Ferrer, Moya Chen, Guillem Cucurull, David Esiobu, Jude Fernandes, Jeremy Fu, Wenyin Fu, Brian Fuller, Cynthia Gao, Vedanuj Goswami, Naman Goyal, Anthony Hartshorn, Saghar Hosseini, Rui Hou, Hakan Inan, Marcin Kardas, Viktor Kerkez, Madian Khabsa, Isabel Kloumann, Artem Korenev, Punit~Singh Koura, Marie{-}Anne Lachaux, Thibaut Lavril, Jenya Lee, Diana Liskovich, Yinghai Lu, Yuning Mao, Xavier Martinet, Todor Mihaylov, Pushkar Mishra, Igor Molybog, Yixin Nie, Andrew Poulton, Jeremy Reizenstein, Rashi Rungta, Kalyan Saladi, Alan Schelten, Ruan Silva, Eric~Michael Smith, Ranjan Subramanian, Xiaoqing~Ellen Tan, Binh Tang, Ross Taylor, Adina Williams, Jian~Xiang Kuan, Puxin Xu, Zheng Yan, Iliyan Zarov, Yuchen Zhang, Angela Fan, Melanie Kambadur, Sharan Narang, Aur{\'{e}}lien Rodriguez, Robert Stojnic, Sergey Edunov,
  and Thomas Scialom. 2023{\natexlab{b}}.
\newblock \href {https://doi.org/10.48550/arXiv.2307.09288} {Llama 2: Open foundation and fine-tuned chat models}.
\newblock \emph{CoRR}, abs/2307.09288.

\bibitem[{Wei et~al.(2022)Wei, Bosma, Zhao, Guu, Yu, Lester, Du, Dai, and Le}]{flan}
Jason Wei, Maarten Bosma, Vincent~Y. Zhao, Kelvin Guu, Adams~Wei Yu, Brian Lester, Nan Du, Andrew~M. Dai, and Quoc~V. Le. 2022.
\newblock \href {https://openreview.net/forum?id=gEZrGCozdqR} {Finetuned language models are zero-shot learners}.
\newblock In \emph{The Tenth International Conference on Learning Representations, {ICLR} 2022}. OpenReview.net.

\bibitem[{Yang et~al.(2024)Yang, Yang, Hui, Zheng, Yu, Zhou, Li, Li, Liu, Huang et~al.}]{yang2024qwen2}
An~Yang, Baosong Yang, Binyuan Hui, Bo~Zheng, Bowen Yu, Chang Zhou, Chengpeng Li, Chengyuan Li, Dayiheng Liu, Fei Huang, et~al. 2024.
\newblock \href {https://arxiv.org/pdf/2407.10671} {Qwen2 technical report}.
\newblock \emph{arXiv preprint arXiv:2407.10671}.

\bibitem[{Yang et~al.(2023)Yang, Chern, Qiu, Neubig, and Liu}]{DBLP:journals/corr/alignment-for-honesty-2023}
Yuqing Yang, Ethan Chern, Xipeng Qiu, Graham Neubig, and Pengfei Liu. 2023.
\newblock \href {https://doi.org/10.48550/ARXIV.2312.07000} {Alignment for honesty}.
\newblock \emph{CoRR}, abs/2312.07000.

\bibitem[{Zhang et~al.(2024)Zhang, Diao, Lin, Fung, Lian, Wang, Chen, Ji, and Zhang}]{zhang-etal-2024-r}
Hanning Zhang, Shizhe Diao, Yong Lin, Yi~Fung, Qing Lian, Xingyao Wang, Yangyi Chen, Heng Ji, and Tong Zhang. 2024.
\newblock \href {https://aclanthology.org/2024.naacl-long.394} {{R}-tuning: Instructing large language models to say {`}{I} don{'}t know{'}}.
\newblock In \emph{Proceedings of the 2024 Conference of the North American Chapter of the Association for Computational Linguistics: Human Language Technologies (Volume 1: Long Papers)}, pages 7113--7139, Mexico City, Mexico. Association for Computational Linguistics.

\bibitem[{Zhang et~al.(2023)Zhang, Aljunied, Gao, Chia, and Bing}]{m3exam}
Wenxuan Zhang, Mahani Aljunied, Chang Gao, Yew~Ken Chia, and Lidong Bing. 2023.
\newblock \href {http://papers.nips.cc/paper\_files/paper/2023/hash/117c5c8622b0d539f74f6d1fb082a2e9-Abstract-Datasets\_and\_Benchmarks.html} {M3exam: {A} multilingual, multimodal, multilevel benchmark for examining large language models}.
\newblock In \emph{Advances in Neural Information Processing Systems 36: Annual Conference on Neural Information Processing Systems 2023, NeurIPS 2023}.

\bibitem[{Zhao et~al.(2024{\natexlab{a}})Zhao, Zhang, Gao, Zhang, Gui, and Huang}]{language-enhancement-llama}
Jun Zhao, Zhihao Zhang, Luhui Gao, Qi~Zhang, Tao Gui, and Xuanjing Huang. 2024{\natexlab{a}}.
\newblock \href {https://doi.org/10.48550/arXiv.2401.01055} {Llama beyond english: An empirical study on language capability transfer}.
\newblock \emph{CoRR}, abs/2401.01055.

\bibitem[{Zhao et~al.(2023)Zhao, Zhou, Li, Tang, Wang, Hou, Min, Zhang, Zhang, Dong, Du, Yang, Chen, Chen, Jiang, Ren, Li, Tang, Liu, Liu, Nie, and Wen}]{llm-survey-zhaoxin}
Wayne~Xin Zhao, Kun Zhou, Junyi Li, Tianyi Tang, Xiaolei Wang, Yupeng Hou, Yingqian Min, Beichen Zhang, Junjie Zhang, Zican Dong, Yifan Du, Chen Yang, Yushuo Chen, Zhipeng Chen, Jinhao Jiang, Ruiyang Ren, Yifan Li, Xinyu Tang, Zikang Liu, Peiyu Liu, Jian{-}Yun Nie, and Ji{-}Rong Wen. 2023.
\newblock \href {https://doi.org/10.48550/arXiv.2303.18223} {A survey of large language models}.
\newblock \emph{CoRR}, abs/2303.18223.

\bibitem[{Zhao et~al.(2024{\natexlab{b}})Zhao, Zhang, Chen, Kawaguchi, and Bing}]{multilingual-analysis}
Yiran Zhao, Wenxuan Zhang, Guizhen Chen, Kenji Kawaguchi, and Lidong Bing. 2024{\natexlab{b}}.
\newblock \href {https://doi.org/10.48550/arXiv.2402.18815} {How do large language models handle multilingualism?}
\newblock \emph{CoRR}, abs/2402.18815.

\bibitem[{Zheng et~al.(2023)Zheng, Chiang, Sheng, Zhuang, Wu, Zhuang, Lin, Li, Li, Xing, Zhang, Gonzalez, and Stoica}]{mtbench}
Lianmin Zheng, Wei{-}Lin Chiang, Ying Sheng, Siyuan Zhuang, Zhanghao Wu, Yonghao Zhuang, Zi~Lin, Zhuohan Li, Dacheng Li, Eric~P. Xing, Hao Zhang, Joseph~E. Gonzalez, and Ion Stoica. 2023.
\newblock \href {http://papers.nips.cc/paper\_files/paper/2023/hash/91f18a1287b398d378ef22505bf41832-Abstract-Datasets\_and\_Benchmarks.html} {Judging llm-as-a-judge with mt-bench and chatbot arena}.
\newblock In \emph{Advances in Neural Information Processing Systems 36: Annual Conference on Neural Information Processing Systems 2023, NeurIPS 2023}.

\end{thebibliography}

\appendix

\end{document}